\documentclass{article}
\usepackage{spconf,amsmath,graphicx}
\usepackage{mathtools} 
\usepackage{url}
\graphicspath{{Figure/}}      
\usepackage{amssymb}   
\usepackage{amsfonts}  
\usepackage{enumitem}
\usepackage{balance}
\usepackage[edges]{forest} 
\usepackage{ragged2e}  
\usepackage{tikz}
\usetikzlibrary{patterns} 

\usepackage{afterpage}       
\usepackage{booktabs,tabularx}
\usepackage{multirow}   
\usepackage{placeins}
\usepackage{float}
\usepackage{capt-of} 
\usepackage{cite}

\makeatletter
\let\orig@thebibliography\thebibliography
\def\thebibliography#1{%
  \orig@thebibliography{#1}%
  \setlength{\itemsep}{0.15ex plus 0.2ex minus 0.1ex}%
  \setlength{\parsep}{0pt}%
  \setlength{\parskip}{0pt}%
}
\makeatother

\newlist{tightenum}{enumerate}{1}
\setlist[tightenum]{%
  label=\arabic*.,      
  leftmargin=*,         
  itemsep=0.25ex plus 0.2ex minus 0.1ex, 
  parsep=0pt,                               
  topsep=0.6ex plus 0.2ex minus 0.1ex,      
  partopsep=0pt
}

\newenvironment{tightmath}{%
  \setlength{\abovedisplayskip}{4pt plus 0.8pt minus 0.6pt}%
  \setlength{\belowdisplayskip}{4pt plus 0.8pt minus 0.6pt}%
  \setlength{\abovedisplayshortskip}{0pt plus 0.5pt}%
  \setlength{\belowdisplayshortskip}{1.5pt plus 0.5pt minus 0.5pt}%
  \setlength{\jot}{1.5pt}
}{}



\title{From Turn-Taking to Synchronous Dialogue: \\
A Survey of Full-Duplex Spoken Language Models}

\name{Yuxuan Chen$^{1*}$ \qquad Haoyuan Yu$^{2}$}
\address{$^{1}$ Jilin University, Changchun, China 
         $^{2}$ Hunan University, Changsha, China \\
         \texttt{yxchen5522@jlu.edu.cn} \quad \texttt{y15352176976@hnu.edu.cn} \\
         }

\begin{document}
\maketitle
\begin{abstract}

True Full-Duplex (TFD) voice communication—enabling simultaneous listening and speaking with natural turn-taking, overlapping speech, and interruptions—represents a critical milestone toward human-like AI interaction. This survey comprehensively reviews Full-Duplex Spoken Language Models (FD-SLMs) in the LLM era. We establish a taxonomy distinguishing Engineered Synchronization (modular architectures) from Learned Synchronization (end-to-end architectures), and unify fragmented evaluation approaches into a framework encompassing Temporal Dynamics, Behavioral Arbitration, Semantic Coherence, and Acoustic Performance. Through comparative analysis of mainstream FD-SLMs, we identify fundamental challenges—synchronous data scarcity, architectural divergence, and evaluation gaps—providing a roadmap for advancing human-AI communication.

For code and further details, please refer to GitHub%
\footnote{\label{repo}\url{https://github.com/elpsykongloo/FD-SLMs}}.

\end{abstract}



\afterpage{%
  \begingroup
  \setlength{\dblfloatsep}{4pt plus 1pt minus 1pt}    
  \setlength{\dbltextfloatsep}{4pt plus 1pt minus 1pt} 
  \setlength{\abovecaptionskip}{1pt plus 0.5pt minus 0.5pt} 
  \setlength{\belowcaptionskip}{0pt}

  \begin{figure*}[!t]
   \centering
   \includegraphics[width=\textwidth]{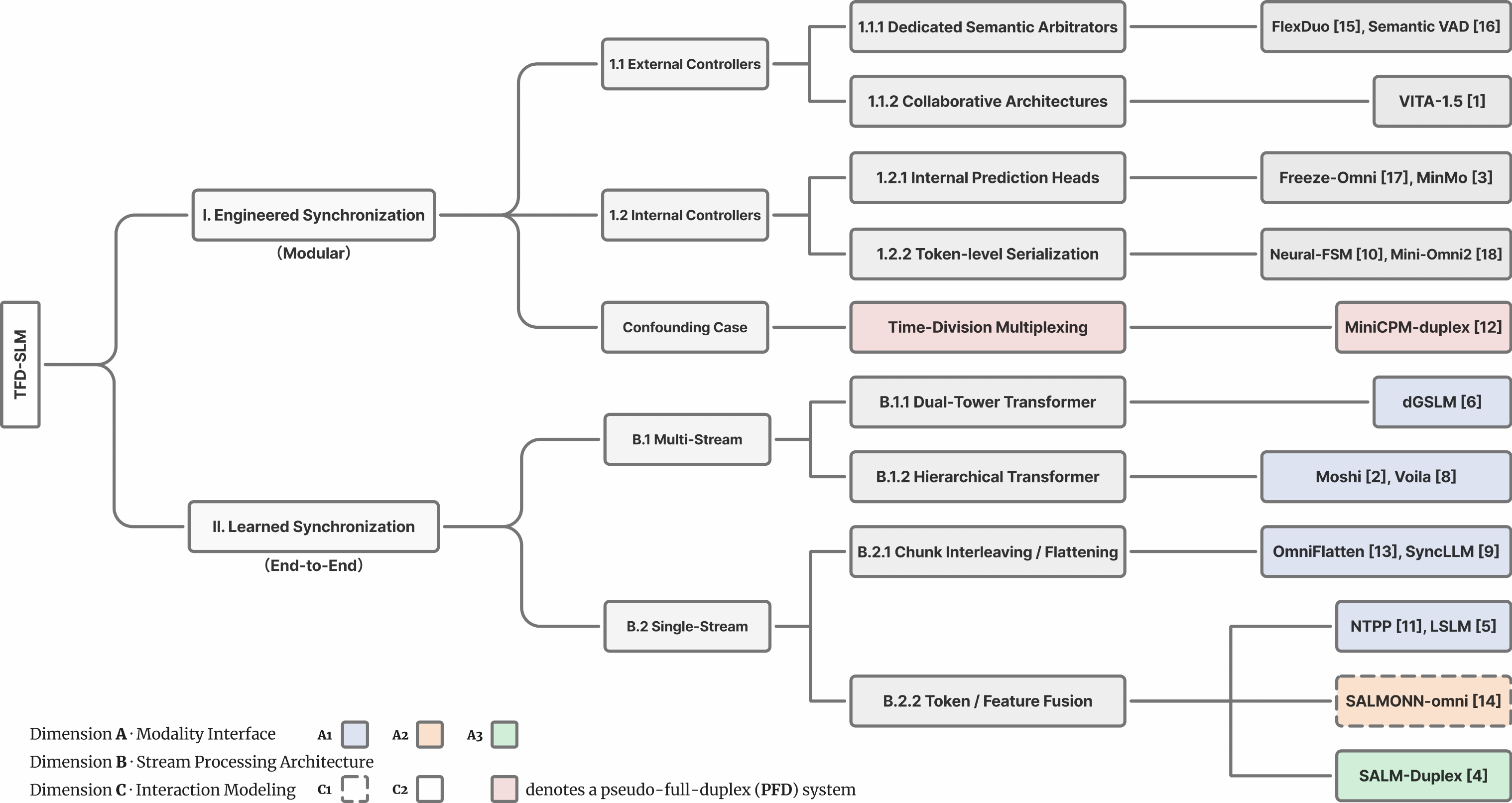}
   \caption{Architectural Taxonomy of Open-source FD\mbox{-}SLMs.}
   \label{fig:fig2}
  \end{figure*}
  \endgroup
}

\afterpage{%
  \afterpage{%
    \afterpage{%
        \begin{table*}[!t]
          \centering
          \caption{Comparative analysis of architectural components in open-source FD-SLMs.}
          \label{tab:arch-compare}
          \setlength{\tabcolsep}{4pt}
          \renewcommand{\arraystretch}{1.15}
          \scriptsize
          \begin{tabular}{l p{4.5cm} p{4.8cm} p{4.2cm}}
            \toprule
            \textbf{Model} & \textbf{Input Perception} & \textbf{Core Processing} & \textbf{Output Synthesis} \\
            \midrule
            \textbf{dGSLM} &
              HuBERT + k-means clustering &
              Two-tower Transformer with cross-attention &
              HiFi-GAN unit vocoder \\
            \textbf{Moshi} &
              Mimi neural codec (RVQ) &
              RQ-Transformer joint autoregression &
              Mimi decoder \\
            \textbf{SyncLLM} &
              HuBERT features &
              Interleaved and predictive synchronization &
              HiFi-GAN vocoder \\
            \textbf{SALMONN-omni} &
              Mamba streaming encoder &
              Dynamic control tokens for stream management &
              CosyVoice2 with fixed-length generation \\
            \textbf{MinMo} &
              SenseVoice-Large + projector &
              Full-Duplex Predictor (FDP) head &
              CosyVoice2 chunk-aware flow-matching \\
            \textbf{FlexDuo} &
              Qwen2-Audio encoder &
              Finite-state machine control &
              External TTS delegation \\
            \textbf{VITA 1.5} &
              Conv + Transformer encoder &
              Dual LLM instances with shared KV cache &
              TiCodec decoder \\
            \bottomrule
          \end{tabular}
        \end{table*}
    }
  }
}

\afterpage{%
  \afterpage{%
    \afterpage{%
     \afterpage{%
      \begingroup
      \setlength{\dblfloatsep}{4pt plus 1pt minus 1pt}     
      \setlength{\dbltextfloatsep}{6pt plus 1pt minus 1pt} 
      \setlength{\abovecaptionskip}{1pt plus 0.5pt minus 0.5pt} 
      \setlength{\belowcaptionskip}{0pt}                   

      \begin{table*}[!t]
        \centering
        \caption{\footnotesize Comprehensive evaluation of representative open-source FD\mbox{-}SLMs across four dimensions.}
        \label{tab:fdslm_eval}
        \setlength{\tabcolsep}{4pt}
        \renewcommand{\arraystretch}{1.15}
        \scriptsize

        \begin{tabular}{l cc ccc cc cc}
          \toprule
          \multirow{2}{*}{\textbf{Model}} &
          \multicolumn{2}{c}{\textbf{Temporal Dynamics}} &
          \multicolumn{3}{c}{\textbf{Behavioral Arbitration}} &
          \multicolumn{2}{c}{\textbf{Semantic Coherence}} &
          \multicolumn{2}{c}{\textbf{Acoustic Performance}} \\
          & \textbf{FTO} ($\downarrow$) & \textbf{SL} ($\downarrow$) &
            \textbf{IRD} ($\downarrow$) & \textbf{ISR} ($\uparrow$) & \textbf{WER} ($\downarrow$) &
            \textbf{PPL} ($\downarrow$) & \textbf{QA Acc} ($\uparrow$) &
            \textbf{N\mbox{-}MOS} ($\uparrow$) & \textbf{M\mbox{-}MOS} ($\uparrow$) \\
          \midrule
          \textbf{Human}         & $\sim$0.20\,s         & $\sim$0.30\,s         & 2.32\,s & 93.69\% & 1.5\%             & 10.2  & 92\%   & 4.92 ($\pm$0.02) & 4.85 ($\pm$0.03) \\
          \textbf{dGSLM}         & 0.33\,s ($\pm$0.12)   & 0.15\,s ($\pm$0.03)   & 1.33\,s & 60.31\% & 25\% ($\pm$3.4)   & 334.4 & 17.2\% & 3.85 ($\pm$0.12) & 1.38 ($\pm$0.10) \\
          \textbf{NTPP}          & 0.30\,s ($\pm$0.15)   & 0.18\,s ($\pm$0.05)   & 1.30\,s & 80.82\% & 7.5\% ($\pm$1.22) & 35    & 55.2\% & 4.15 ($\pm$0.06) & 3.95 ($\pm$0.04) \\
          \textbf{Moshi}         & 2.22\,s ($\pm$0.70)   & 0.75\,s ($\pm$0.10)   & 1.44\,s & 77.73\% & 5.20\% ($\pm$0.13)& 59.3  & 33.8\% & 3.90 ($\pm$0.07) & 3.75 ($\pm$0.06) \\
          \textbf{SALMONN-omni}  & 0.38\,s ($\pm$0.10)   & 0.25\,s ($\pm$0.08)   & 1.38\,s & 85.6\%  & 8.40\% ($\pm$0.20)& 21.1  & 61\%   & 3.85 ($\pm$0.10) & 3.95 ($\pm$0.15) \\
          \textbf{VITA-1.5}      & 2.10\,s ($\pm$0.65)   & 0.12\,s ($\pm$0.05)   & 9.49\,s & 78.53\% & 5.45\% ($\pm$0.10)& 26.8  & 50.5\% & 4.00 ($\pm$0.08) & 4.10 ($\pm$0.10) \\
          \textbf{Freeze-Omni}   & $-0.40$\,s ($\pm$0.05)& 1.11\,s ($\pm$0.17)   & 9.25\,s & 54.97\% & 7.30\% ($\pm$0.05)& 30.2  & 56.9\% & 3.80 ($\pm$0.10) & 3.90 ($\pm$0.07) \\
          \bottomrule
        \end{tabular}
      \end{table*}
      \endgroup
     }
    }
  }
}


\section{Introduction}
Contemporary SLMs fundamentally lack simultaneous listening and speaking capabilities essential for natural conversation. While LLMs have revolutionized language understanding \cite{GPT-4o,Step-Audio-2}, their spoken dialogue implementations remain constrained by sequential listen-think-speak cycles. Current systems achieve only pseudo-full-duplex (PFD) behavior through time-division multiplexing, failing to match human conversational dynamics \cite{Universals,Timing} characterized by natural turn-taking behaviors illustrated in Fig.~\ref{fig:fig1}.

FD-SLMs transform this paradigm from sequential to parallel cognitive architectures. Unlike PFD systems that alternate between listening and speaking, FD-SLMs enable simultaneous encoding and generation within unified processing cycles, supporting natural conversational events including interruptions, backchanneling, and adaptive turn-taking through bidirectional information flow.

\FloatBarrier 

\begingroup
\setlength{\textfloatsep}{0pt}
\setlength{\intextsep}{0pt}
\setlength{\abovecaptionskip}{0.5pt plus 0.5pt minus 0.5pt}
\setlength{\belowcaptionskip}{0pt}
\begin{figure}[!t]
  \centering
  \includegraphics[width=\columnwidth]{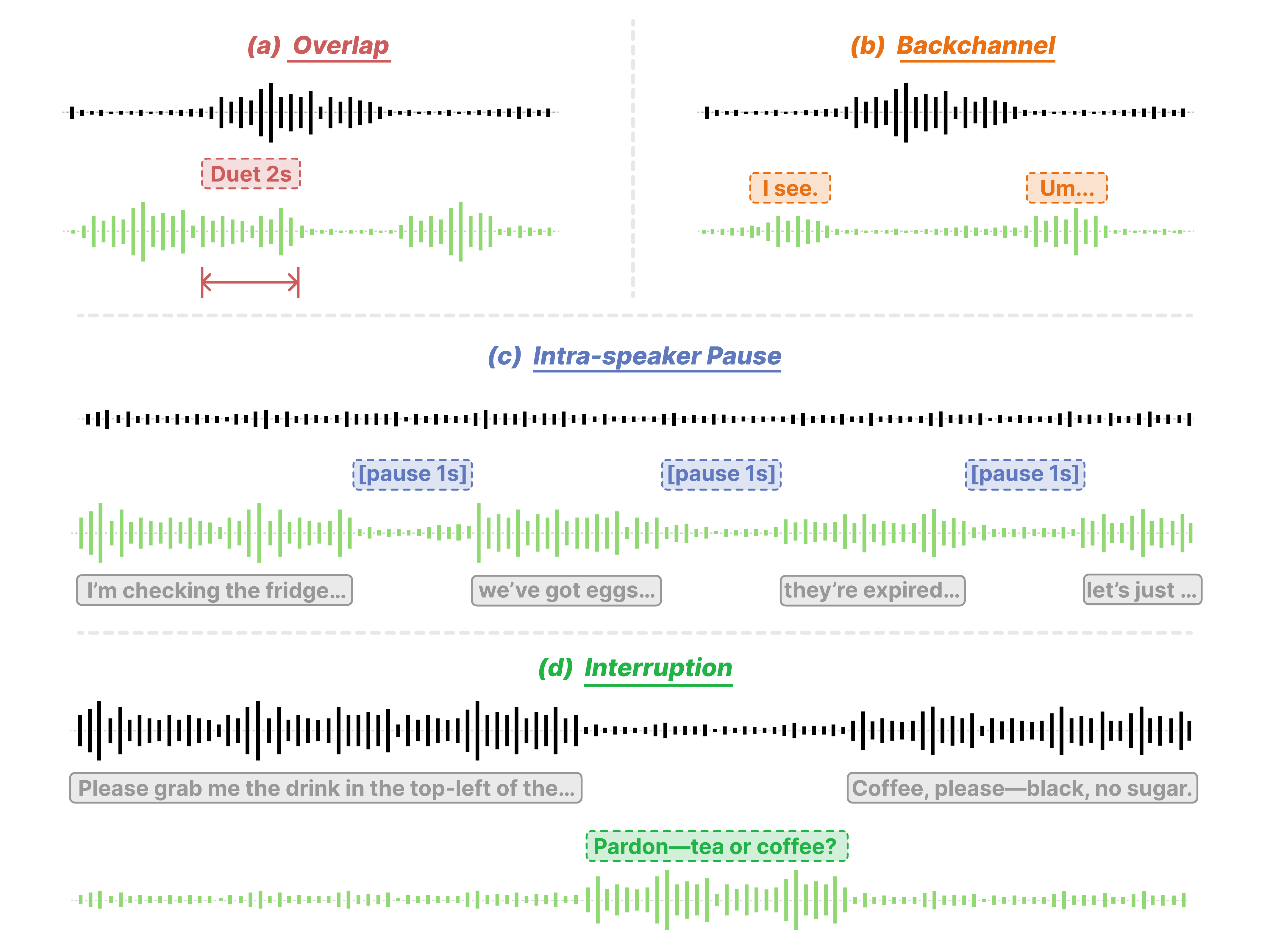}
  \caption{
    Natural conversations contain turn-taking events:
    (a) Overlap, (b) Backchannel, (c) Pause, and (d) Interruption.
  }
  \label{fig:fig1}
\end{figure}
\endgroup

Early systems demonstrated incremental processing \cite{Finite-State} and finite-state control \cite{Turn-taking}, achieving responsiveness without semantic flexibility. LLM integration yielded engineered synchronization \cite{FlexDuo,Neural-FSM,OmniFlatten} and end-to-end architectures. Following dGSLM's emergent turn-taking discovery \cite{dGSLM}, recent advances include hierarchical multi-stream processing \cite{Moshi}, next-token-pair prediction (NTPP)\cite{NTPP}, and continuous-discrete alignment \cite{SALMONN-omni}.

Despite these advances, existing surveys \cite{Landscape,ResentAdvance} treat full-duplex as implementation detail rather than fundamental requirement, lacking systematic FD-SLM design analysis. Evaluation also remains fragmented\cite{FD-Bench,FD-1.0,FD-1.5}.

\vspace{0.3ex}
This paper makes the following primary contributions:
\vspace{0.2ex}
\begin{tightenum} [leftmargin=*]
  \item \textbf{Formal duplex characterization:} Mathematical definitions rigorously distinguish half-duplex, pseudo-full-duplex, and true full-duplex systems, exposing computational requirements for cognitive parallelism.
  \item \textbf{Architectural taxonomy:} Systematic categorization reveals the design space along synchronization strategy, state management, and training paradigm axes, identifying trade-offs and unexplored opportunities.
  \item \textbf{Systematic evaluation and analysis:} We compare streaming architectures across seven FD-SLMs, identify critical data bottlenecks, and establish a four-pillar benchmarking taxonomy revealing the fundamental latency-quality trade-off that constrains current systems.
\end{tightenum}

\section{Formalization}
\begin{tightmath}

FD\mbox{-}SLMs realize cognitive parallelism by concurrently encoding inputs and decoding outputs, enabling real-time output adaptation. Let agent $\mathcal{A}$ interact with environment $\mathcal{E}$. Because direct modeling of continuous audio $X(t)$ is intractable, a discretizer $\mathcal{T}$ yields aligned sequences $S^\mathcal{E}=(e_1,\ldots,e_T)$ and $S^\mathcal{A}=(a_1,\ldots,a_T)$, aligning $e_t$ with $a_t$ for synchronous interaction.

\vspace{-0.65\baselineskip}
\subsection{Joint Probability Perspective}

\vspace{-0.3\baselineskip}
The interaction modeling paradigm models joint distribution $P(S^\mathcal{E},S^\mathcal{A})$:

\vspace{-0.9\baselineskip}
\begin{equation}
P(S^\mathcal{E}, S^\mathcal{A})=\prod_{t=1}^{T} P\!\left(e_t,a_t \mid S^\mathcal{E}_{<t}, S^\mathcal{A}_{<t}\right).
\label{eq:joint}
\end{equation}

This underlies NTPP\cite{NTPP}, simultaneously predicting $(e_t,a_t)$ pairs in decoder-only transformers:
\begin{equation}
\resizebox{.91\columnwidth}{!}{$
\mathcal{L}_{\text{NTPP}}(\theta)=\mathbb{E}_{(S^\mathcal{E},S^\mathcal{A})}\!\left[\sum_{t=1}^{T}\log P\!\left(e_t,a_t \mid S^\mathcal{E}_{<t},S^\mathcal{A}_{<t};\theta\right)\right]
$}
\label{eq:ntpp}
\end{equation}

Earlier approaches\cite{dGSLM} approximate through conditional independence with cross-attention, optimizing summed conditional log-likelihoods rather than true joint likelihood.

\vspace{-0.65\baselineskip}
\subsection{Conditional Probability Perspective}
\vspace{-0.2\baselineskip}
For interactive agents, the objective becomes modeling $P(S^\mathcal{A}\mid S^\mathcal{E})$:
\vspace{-0.7\baselineskip}
\begin{equation}
a_t \sim P\!\left(a_t \mid S^\mathcal{E}_{\le t},\, S^\mathcal{A}_{<t};\,\theta\right).
\label{eq:cond}
\end{equation}

\begingroup
\makeatletter
\renewcommand\paragraph{\@startsection{paragraph}{4}{\z@}%
  {2pt \@plus 1pt \@minus 1pt}
  {-0.5em}
  {\normalfont\normalsize\bfseries}}
\makeatother

\paragraph*{Concurrency.}
Computing $a_t$ while ingesting $e_{t+1},\ldots$ requires parallel encoding--decoding \cite{Moshi,LSLM}.

\paragraph*{Real-time constraint.}
$\mathrm{Time}(\mathrm{Compute}(a_t))< 200\,\mathrm{ms}$ \cite{Universals}.

\paragraph*{Self-conditioning.}
Dependence on $S^\mathcal{A}_{<t}$ ensures coherence and enables echo cancellation \cite{SALMONN-omni}.
\endgroup

The training objective:
\begin{equation}
\resizebox{.91\columnwidth}{!}{$
\mathcal{L}_{\text{Cond}}(\theta)
= \mathbb{E}_{(S^\mathcal{E},S^\mathcal{A})}\!\left[\sum_{t=1}^{T}
\log P\!\left(a_t \mid S^\mathcal{E}_{\le t},\, S^\mathcal{A}_{<t};\,\theta\right)\right]
$}
\label{eq:cond-loss}
\end{equation}

Training on synchronous data with either objective enables turn-taking dynamics to emerge without supervision.

\vspace{-0.65\baselineskip}
\subsection{Hierarchical and Predictive Mechanisms}
\begingroup
\makeatletter
\renewcommand\paragraph{\@startsection{paragraph}{4}{\z@}%
  {2pt \@plus 1pt \@minus 1pt}%
  {-0.5em}%
  {\normalfont\normalsize\bfseries}}
\makeatother

\paragraph*{Hierarchical Generation.}
Systems like Moshi \cite{Moshi} leverage text representations $T^\mathcal{A}$:
\vspace{-0.1\baselineskip}
\begin{equation}
\resizebox{.91\columnwidth}{!}{$
P(S^\mathcal{A} \mid S^\mathcal{E})
= \int P\!\left(S^\mathcal{A}\mid T^\mathcal{A}, S^\mathcal{E}\right)\,
     P\!\left(T^\mathcal{A}\mid S^\mathcal{E}\right)\, dT^\mathcal{A}
$}
\label{eq:hier}
\end{equation}

\vspace{-0.1\baselineskip}
A two-stage process generates text tokens ("inner monologue") then audio tokens, merging text-based reasoning with full-duplex capabilities.

\vspace{-0.1\baselineskip}
\paragraph*{Predictive Synchronization.}
SyncLLM \cite{SyncLLM} predicts upcoming user segments to minimize latency:
\vspace{-0.1\baselineskip}
\begin{equation}
\resizebox{.91\columnwidth}{!}{$
\hat e_{t+1} \sim P\!\left(\cdot \mid S^\mathcal{E}_{\le t}, S^\mathcal{A}_{\le t}\right),\quad
a_{t+1} \sim P\!\left(\cdot \mid S^\mathcal{E}_{\le t}, \hat e_{t+1}, S^\mathcal{A}_{\le t}\right)
$}
\label{eq:sync}
\end{equation}

\endgroup
\end{tightmath}

\section{Taxonomy}
Cognitive parallelism, enabling simultaneous speech encoding and output decoding, requires departing from sequential Transformer architectures. Figure~\ref{fig:fig2} shows current approaches following two paradigms: \textbf{engineered synchronization} via modular architectures and \textbf{learned synchronization} through end-to-end systems.

\vspace{-0.4\baselineskip}
\subsection{Engineered Synchronization}
Modular approaches enhance dialogue engines with specialized components, eliminating retraining through explicit state arbitration. The duplex controller—a neural FSM—extends beyond acoustic VAD to perform semantic arbitration, differentiating interruptions from backchannels and noise.

\vspace{-0.3\baselineskip}
\paragraph*{External controllers.}
External controllers maintain independence from the core engine. FlexDuo introduces a ternary FSM with an idle state for selective attention~\cite{FlexDuo}. Semantic VAD uses lightweight ($\sim$0.5B) models analyzing ASR outputs to minimize computational load~\cite{SemanticVAD}. VITA\mbox{-}1.5 employs dual instances that swap roles upon interruption detection, trading computational cost for latency~\cite{VITA-1.5}.

\vspace{-0.7\baselineskip}
\paragraph*{Internal controllers.}
Internal controllers embed control logic within the engine architecture. Freeze\mbox{-}Omni performs chunk-wise state prediction on frozen LLMs~\cite{Freeze-Omni}; MinMo’s Full Duplex Predictor reads embeddings for turn-yielding decisions~\cite{Mini-Omni2}. Neural\mbox{-}FSM extends vocabularies with FSM tokens enabling autonomous state management through next-token prediction~\cite{Neural-FSM}. Mini\mbox{-}Omni2 implements command-based interruption via semantic state tokens~\cite{Mini-Omni2}.

\vspace{-0.4\baselineskip}
\subsection{Learned Synchronization}
End-to-end architectures natively process bidirectional audio streams. Following dGSLM’s demonstration of emergent turn-taking from raw audio~\cite{dGSLM}, these systems make full-duplex capabilities intrinsic. The challenge lies in reconciling Transformer sequentiality with conversational parallelism.

\vspace{-0.3\baselineskip}
\paragraph*{Modal interfaces.}
Modal interfaces vary in representation. Codec-based approaches~\cite{dGSLM,Moshi,SyncLLM,NTPP,Voila} discretize audio into tokens despite sequence elongation. SALMONN\mbox{-}omni directly processes continuous embeddings~\cite{SALMONN-omni}. SALM\mbox{-}Duplex combines continuous inputs with discrete outputs for an accuracy–latency tradeoff~\cite{SALM}.

\vspace{-0.7\baselineskip}
\paragraph*{Stream processing.}
Stream processing follows multi-stream or single-stream paradigms. Multi-stream approaches like dual-tower architectures use cross-attention for synchronization~\cite{dGSLM}, while Moshi's RQ-Transformer jointly models user/agent audio and internal monologue~\cite{Moshi}. Single-stream methods serialize inputs for standard decoders: SyncLLM interleaves chunks with synchronization tokens~\cite{SyncLLM}, NTPP uses pairwise causal masking~\cite{NTPP}, and LSLM/SALM-Duplex explore varying fusion depths~\cite{LSLM,SALM}.

\vspace{-0.7\baselineskip}
\paragraph*{Interaction modeling.}
Interaction modeling predominantly employs implicit dynamics where models control turn-taking through silence or audible token generation without explicit supervision~\cite{dGSLM,Moshi,SyncLLM,NTPP,Voila}. In contrast, SALMONN-omni's Dynamic Thinking mechanism~\cite{SALMONN-omni} generates control tokens for explicit state management, positioning the LLM as the duplex predictor within an end-to-end framework.

\section{Evaluation}
FD-SLMs demand coordinated assessment across three interdependent axes: streaming architectures enabling real-time interaction, conversational training data, and comprehensive benchmarking methodologies.

\vspace{-0.4\baselineskip}
\subsection{Architectural Components}

\vspace{-0.4\baselineskip}
FD\mbox{-}SLMs require specialized streaming architectures achieving sub\mbox{-}200\,ms latency for natural turn-taking \cite{Landscape,ResentAdvance}. Table~\ref{tab:arch-compare} summarizes strategies across three critical stages.

\vspace{-0.3\baselineskip}
\paragraph*{\textbf{Input Perception.}}
Continuous encoding with minimal lookahead is essential. While conventional encoders need causal adaptation, purpose-built streaming encoders operate natively\cite{SALMONN-omni,Emformer,Zipformer}. Discrete paradigms employ strictly causal/near-zero-lookahead neural codecs\cite{Moshi,NTPP}; tokenizer chunk granularity fundamentally bounds perception latency \cite{NTPP,3Ways,RVQGAN-HiFi,WavTokenizer}.

\vspace{-0.7\baselineskip}
\paragraph*{\textbf{Core Processing.}}
Concurrent streams are synchronized via cross-attention \cite{dGSLM}, joint autoregression \cite{Moshi}, predictive synchronization \cite{SyncLLM}, or explicit control mechanisms \cite{Mini-Omni2,FlexDuo}. A 100–200\,ms ``cognitive clock’’ sets perception–reaction granularity \cite{SyncLLM,Moshi,OmniFlatten}. KV\mbox{-}cache efficiency directly affects sustained responsiveness \cite{NTPP,Connectors}.

\vspace{-0.7\baselineskip}
\paragraph*{\textbf{Output Synthesis.}}
Discrete models reuse codec decoders for minimal latency \cite{SyncSpeech,LLMVoX}. Continuous pipelines employ chunk-aware flow-matching \cite{MinMo}, fixed-length interleaved generation \cite{SALMONN-omni}, or tightly coupled LLM–vocoder stacks\cite{VITA-1.5}.

\vspace{-0.4\baselineskip}
\subsection{Training Data}

Data scarcity remains critical: FD-SLMs require synchronized multi-channel spontaneous dialogue, unavailable in monologue corpora. Current training uses limited datasets \cite{dGSLM,Moshi,NTPP} constraining diversity (see Table~\ref{tab:datasets} for examples; full listings in our repository).

\FloatBarrier
\begingroup

\setlength{\textfloatsep}{2pt plus 0.5pt minus 0.5pt}   
\setlength{\intextsep}{1.5pt plus 0.5pt minus 0.5pt}    
\setlength{\abovecaptionskip}{0pt}                      
\setlength{\belowcaptionskip}{1pt plus 0.5pt minus 0.5pt} 
\setlength{\tabcolsep}{2pt}                             
\renewcommand{\arraystretch}{1.06}                      

\begin{table}[H] 
  \caption{Publicly Available Datasets for FD\mbox{-}SLM}
  \label{tab:datasets}
  \centering
  \footnotesize
  \begin{tabularx}{\columnwidth}{l c c c c}
    \toprule
    \textbf{Dataset} & \textbf{Lang} & \textbf{Scene} & \textbf{Channels} & \textbf{Hours} \\
    \midrule
    AMI Meeting Corpus           & EN    & meeting   & 8 & 100 \\
    ICSI Meeting Corpus          & EN    & meeting   & 6 & 70  \\
    LibriCSS                     & EN    & meeting   & 7 & 10        \\
    Fisher English               & EN    & phone     & 2 & 1{,}960 \\
    SEAME (Mandarin--English CS) & EN+ZH & interview & 2 & 192 \\
    HKUST Mandarin Telephone     & ZH    & phone     & 2 & 149 \\
    \bottomrule
  \end{tabularx}
\end{table}

\endgroup
\FloatBarrier

Synthetic TTS generation \cite{SyncLLM} fails to capture prosodic entrainment and overlap dynamics, limiting generalization. Progress requires end-to-end conversational synthesis and advanced source separation for single-channel data.

\vspace{-0.4\baselineskip}
\subsection{Benchmarking Framework}

Conventional metrics built for half-duplex systems \cite{GPT-4o,UTMOS} fail to capture real-time FD behaviors: when models speak, how they intervene, and conversational floor arbitration \cite{Turn-taking}.

Historical fragmentation through model-specific metrics \cite{Moshi,dGSLM,NTPP,Neural-FSM} prevented systematic comparison. Recent standardization efforts \cite{FD-1.0,FD-1.5,FD-Bench} enable reproducible evaluation via our four-pillar taxonomy (Fig.~\ref{fig:fig3}).

Table~\ref{tab:fdslm_eval} reveals critical gaps: while acoustic quality approaches human levels, temporal dynamics vary widely, behavioral arbitration underperforms (ISR: 54-86\% vs. 94\% human), and semantic coherence trades off against responsiveness—demonstrating that human-parity FD requires paradigmatic architectural advances.

{
  \setlength{\abovecaptionskip}{3pt}   
  \setlength{\belowcaptionskip}{0pt}   
  \setlength{\textfloatsep}{4pt plus 1pt minus 1pt} 

  \begin{figure}[!t] 
    \centering
    \includegraphics[width=1\columnwidth,height=.33\textheight,keepaspectratio]{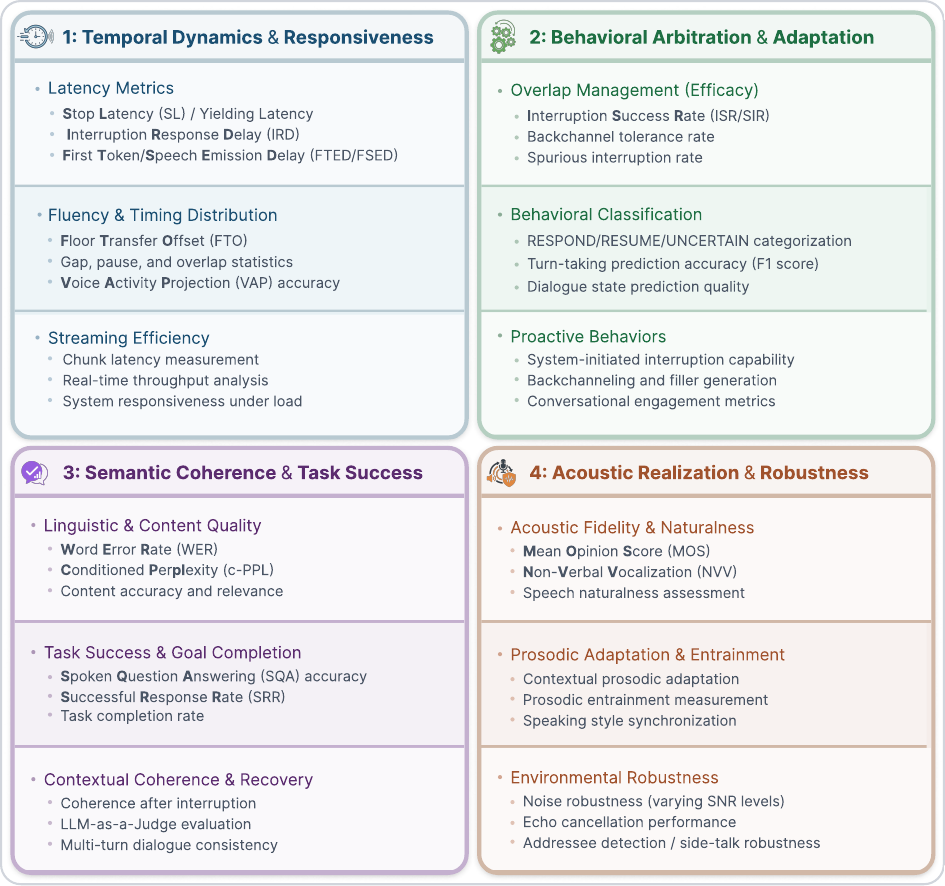}
    \caption{Four\mbox{-}Pillar Taxonomy of Benchmarking FD\mbox{-}SLMs.}
    \label{fig:fig3}
  \end{figure}

  \vspace*{-3pt}
  \FloatBarrier 
}

\section{CONCLUSION}

FD-SLMs mark a paradigm shift from turn-based to synchronous dialogue. Through cognitive concurrency formalization and our taxonomy distinguishing Engineered from Learned Synchronization, we clarify fundamental design trade-offs. Our four-pillar evaluation reveals that while acoustic quality approaches human levels, critical gaps persist: inconsistent temporal dynamics, suboptimal behavioral arbitration, and inverse latency-coherence correlation.

Progress requires addressing interconnected challenges. Architectural fragmentation prevents scalable designs aligned with LLM scaling laws. Data scarcity—particularly synchronized multi-channel recordings and non-English resources \cite{JapMoshi}—constrains learning. Current evaluation lacks proactive behavior metrics \cite{Voila}, while ultra-low latency introduces safety risks requiring real-time filtering.

Advancing FD-SLMs demands architectural convergence, synthetic data capturing authentic dynamics, comprehensive behavioral evaluation, and robust safety mechanisms. Only through coordinated efforts can we achieve truly human-like conversational AI that is responsive, scalable, and ethically deployable.

\clearpage   
\nocite{*}
\bibliographystyle{IEEEbib}
\bibliography{refs}

\begin{thebibliography}{10}

\bibitem{GPT-4o}
OpenAI et~al.,
\newblock ``{GPT-4o} system card,''
\newblock 2024.

\bibitem{Step-Audio-2}
Boyong Wu et~al.,
\newblock ``Step-audio 2 technical report,''
\newblock 2025.

\bibitem{Universals}
Tanya Stivers et~al.,
\newblock ``Universals and cultural variation in turn-taking in conversation,''
\newblock {\em PNAS}, 2009.

\bibitem{Timing}
Stephen~C. Levinson and Francisco Torreira,
\newblock ``Timing in turn-taking and its implications for processing models of
  language,''
\newblock {\em Frontiers in Psychology}, 2015.

\bibitem{Finite-State}
Antoine Raux and Maxine Eskenazi,
\newblock ``A finite-state turn-taking model for spoken dialog systems,''
\newblock in {\em Proc. NAACL-HLT}, 2009.

\bibitem{Turn-taking}
Gabriel Skantze,
\newblock ``Turn-taking in conversational systems and human-robot interaction:
  A review,''
\newblock {\em Computer Speech \& Language}, 2021.

\bibitem{FlexDuo}
Ziyang Liao et~al.,
\newblock ``Flexduo: A pluggable system for enhancing spoken dialogue models
  with full-duplex capabilities,''
\newblock 2025.

\bibitem{Neural-FSM}
Zheng Wang et~al.,
\newblock ``Neural-{FSM}: A full-duplex speech dialogue scheme based on large
  language model,''
\newblock in {\em Proc. NeurIPS}, 2024.

\bibitem{OmniFlatten}
Jun Zhang et~al.,
\newblock ``Omniflatten: A unified framework for spoken language model via
  progressive flattening,''
\newblock in {\em Proc. ACL}, 2025.

\bibitem{dGSLM}
Anh-Duy Nguyen et~al.,
\newblock ``Generative spoken dialogue language modeling,''
\newblock {\em TACL}, 2023.

\bibitem{Moshi}
Alexandre D{\'e}fossez et~al.,
\newblock ``Moshi: a speech-text foundation model for real-time dialogue,''
\newblock 2024.

\bibitem{NTPP}
Qichao Wang et~al.,
\newblock ``{NTPP}: Generative speech language modeling for dual-channel spoken
  dialogue via next-token-pair prediction,''
\newblock 2025.

\bibitem{SALMONN-omni}
Dong Yu et~al.,
\newblock ``Salmonn-omni: A codec-free {LLM} for full-duplex speech
  understanding and generation,''
\newblock 2025.

\bibitem{Landscape}
Siddhant Arora et~al.,
\newblock ``On the landscape of spoken language models: A comprehensive
  survey,''
\newblock 2025.

\bibitem{ResentAdvance}
Jia Cui et~al.,
\newblock ``Recent advances in speech language models: A survey,''
\newblock 2025.

\bibitem{FD-Bench}
Yizhou Peng et~al.,
\newblock ``Fd-bench: A full-duplex benchmarking pipeline designed for full
  duplex spoken dialogue systems,''
\newblock 2025.

\bibitem{FD-1.0}
Guan-Ting Lin et~al.,
\newblock ``Full-duplex-bench: A benchmark to evaluate full-duplex spoken
  dialogue models on turn-taking capabilities,''
\newblock 2025.

\bibitem{FD-1.5}
Guan-Ting Lin et~al.,
\newblock ``Full-duplex-bench v1.5: Evaluating overlap handling for full-duplex
  speech models,''
\newblock 2025.

\bibitem{LSLM}
Ziyang Ma et~al.,
\newblock ``Language model can listen while speaking,''
\newblock 2024.

\bibitem{SyncLLM}
Aditya Veluri et~al.,
\newblock ``Beyond turn-based interfaces: Synchronous {LLM}s as full-duplex
  dialogue agents,''
\newblock 2024.

\bibitem{SemanticVAD}
Mohan Shi et~al.,
\newblock ``Semantic {VAD}: Low-latency voice activity detection for speech
  interaction,''
\newblock 2023.

\bibitem{VITA-1.5}
Chaoyou Fu et~al.,
\newblock ``Vita-1.5: Towards {GPT-4o} level real-time vision and speech
  interaction,''
\newblock 2025.

\bibitem{Freeze-Omni}
Xiong Wang et~al.,
\newblock ``Freeze-omni: A smart and low latency speech-to-speech dialogue
  model with frozen {LLM},''
\newblock 2024.

\bibitem{Mini-Omni2}
Zhifei Xie et~al.,
\newblock ``Mini-omni2: Towards open-source {GPT-4o} with vision, speech and
  duplex capabilities,''
\newblock 2024.

\bibitem{Voila}
Yemin Shi et~al.,
\newblock ``Voila: A sophisticated, synchronous, and swift spoken language
  model,''
\newblock 2025.

\bibitem{SALM}
Ke~Hu et~al.,
\newblock ``Salm-duplex: Efficient and direct duplex modeling for
  speech-to-speech language model,''
\newblock 2025.

\bibitem{Emformer}
Yangyang Shi et~al.,
\newblock ``Emformer: Efficient memory transformer based acoustic model for low
  latency streaming speech recognition,''
\newblock in {\em Proc. ICASSP}, 2021.

\bibitem{Zipformer}
Zengwei Yao et~al.,
\newblock ``Zipformer: A faster and better encoder for automatic speech
  recognition,''
\newblock in {\em Proc. ICLR}, 2024.

\bibitem{3Ways}
Pooneh Mousavi et~al.,
\newblock ``Discrete audio tokens: More than a survey!,''
\newblock 2025.

\bibitem{RVQGAN-HiFi}
Rithesh Kumar et~al.,
\newblock ``High-fidelity audio compression with improved {RVQGAN},''
\newblock in {\em Proc. NeurIPS}, 2023.

\bibitem{WavTokenizer}
Yuanzhe Xu et~al.,
\newblock ``Wavtokenizer: A novel general-purpose audio-to-token converter,''
\newblock in {\em Proc. ICLR}, 2025.

\bibitem{Connectors}
Zhen Li et~al.,
\newblock ``Connector-s: A survey of connectors in multimodal large language
  models,''
\newblock 2025.

\bibitem{SyncSpeech}
Zhengyan Sheng et~al.,
\newblock ``Syncspeech: Low-latency and efficient dual-stream text-to-speech
  based on temporal masked transformer,''
\newblock 2025.

\bibitem{LLMVoX}
Sambal Shikhar et~al.,
\newblock ``{LLMVoX}: A zero-shot, personalized, and streaming speech synthesis
  leveraging large language models,''
\newblock in {\em Findings of ACL}, 2025.

\bibitem{MinMo}
Qian Chen et~al.,
\newblock ``Minmo: A multimodal large language model for seamless voice
  interaction,''
\newblock 2025.

\bibitem{UTMOS}
Takaaki Saeki et~al.,
\newblock ``{UTMOS}: {UTokyo}{-}sarulab {MOS} prediction system for voice
  conversion challenge 2022,''
\newblock in {\em Proc. SSW}, 2022.

\bibitem{JapMoshi}
Shintaro Ohashi et~al.,
\newblock ``Towards a japanese full-duplex spoken dialogue system: Data
  collection, modeling, and evaluation,''
\newblock 2025.

\bibitem{MiniCPM-duplex}
OpenBMB et~al.,
\newblock ``Minicpm-llama3-v 2.5: An 8b-scale multimodal {LLM},''
\newblock 2024.

\bibitem{Understanding}
Jing Peng et~al.,
\newblock ``A survey on speech large language models for understanding,''
\newblock 2024.

\bibitem{WavChat}
Jinglin Chen et~al.,
\newblock ``Wavchat: A survey of spoken dialogue models,''
\newblock 2024.

\bibitem{DialSystem}
Guillermo Castillo-L{\'o}pez et~al.,
\newblock ``A survey of recent advances on turn-taking modeling in spoken
  dialogue systems,''
\newblock in {\em Proc. IWSDS}, 2025.

\bibitem{EnCodec}
Alexandre D{\'e}fossez et~al.,
\newblock ``High-fidelity neural audio compression,''
\newblock {\em Trans. Mach. Learn. Res.}, 2023.

\bibitem{Vocos}
Hubert Siuzdak,
\newblock ``Vocos: Closing the gap between time-domain and fourier-based neural
  vocoders for high-quality audio synthesis,''
\newblock in {\em Proc. ICLR}, 2024.

\end{thebibliography}

\end{document}